%% file: root.tex
\documentclass[letterpaper, 10 pt, conference]{ieeeconf}
\IEEEoverridecommandlockouts        
\usepackage{amsmath}
\usepackage{amssymb}
\usepackage{multirow}
\usepackage{acronym}
\usepackage{bbm}
\usepackage{graphicx}
\usepackage{hyperref}

\title{\LARGE \bf
DirectTracker: 3D Multi-Object Tracking Using Direct Image Alignment and Photometric Bundle Adjustment
}

\author{Mariia Gladkova, Nikita Korobov, Nikolaus Demmel, Aljo\v{s}a O\v{s}ep, Laura Leal-Taix\'e and Daniel Cremers%
\thanks{All authors are with Technical University of Munich (TUM). Project page: \url{https://vision.in.tum.de/research/vslam/directtracker}}%
}

\acrodef{vo}[VO]{visual odometry}
\acrodef{slam}[SLAM]{simultaneous localization and mapping}
\acrodef{ba}[BA]{bundle adjustment}
\acrodef{dof}[DoF]{degrees of freedom}
\acrodef{dia}[DIA]{direct image alignment}
\acrodef{iou}[IoU]{intersection over union}
\acrodef{sfm}[SfM]{structure from motion}
\acrodef{lssd}[LSSD]{locally-scaled sum of squared differences}
\acrodef{mot}[MOT]{multi-object tracking}
\acrodef{giou}[GIoU]{generalized intersection over union}
\acrodef{hota}[HOTA]{higher-order tracking accuracy}
\acrodef{mota}[MOTA]{multi-object tracking accuracy}
\acrodef{motp}[MOTP]{multi-object tracking precision}
\acrodef{fp}[FP]{false positives}
\acrodef{fn}[FN]{false negatives}
\acrodef{idsw}[IDSW]{id switches}

\begin{document}

\maketitle
\thispagestyle{empty}
\pagestyle{empty}

\begin{abstract}
Direct methods have shown excellent performance in the applications of visual odometry and SLAM. In this work we propose to leverage their effectiveness for the task of 3D multi-object tracking. To this end, we propose \textit{DirectTracker}, a framework that effectively combines direct image alignment for the short-term tracking and sliding-window photometric bundle adjustment for 3D object detection. Object proposals are estimated based on the sparse sliding-window pointcloud and further refined using an optimization-based cost function that carefully combines 3D and 2D cues to ensure consistency in image and world space. We propose to evaluate 3D tracking using the recently introduced higher-order tracking accuracy (HOTA) metric and the generalized intersection over union similarity measure to mitigate the limitations of the conventional use of intersection over union for the evaluation of vision-based trackers. We perform evaluation on the KITTI Tracking benchmark for the Car class and show competitive performance in tracking objects both in 2D and 3D.
\end{abstract}

\section{Introduction}
3D dynamic scene understanding is an essential task for various applications in mobile robotics, autonomous driving and augmented reality. In the process of planning the next manoeuvre an autonomous system requires to accurately localize other traffic participants and consistently track them over time.

A significant amount of work has been devoted to development of accurate 3D detectors \cite{li2019stereo, qi2018frustum, chen2020dsgn, sun2020disprcnn} and design of robust 3D trackers \cite{weng2020gnn3dmot, weng2020ab3dmot, kim2021eagermot} that are capable of assigning consistent ids to 3D detections corresponding to the same physical objects. To a large extent the performance of such tracking methods depends on the accuracy of utilized 3D detectors. At the same time many powerful direct \ac{vo} and \ac{slam} systems \cite{engel2014lsd, engel2017direct} have recently emerged showing state-of-the-art performance in 6 \ac{dof} camera localization and 3D scene reconstruction given just an image sequence as input. While object-based \ac{slam} methods have been proposed before, they are limited to static scenes \cite{salas2013slam++}, oriented towards precise camera tracking \cite{bescos2018dynaslam} or focused solely on the accuracy of 3D object proposals \cite{yang2019cubeslam}. Our work is intended to bridge the gap between two communities. In particular, we offer a \ac{mot} method that tracks and localizes dynamic objects by means of techniques proposed in the \ac{slam} community without any reliance on external 3D detections.

\begin{figure}
    \centering
    \includegraphics[width=0.5\textwidth]{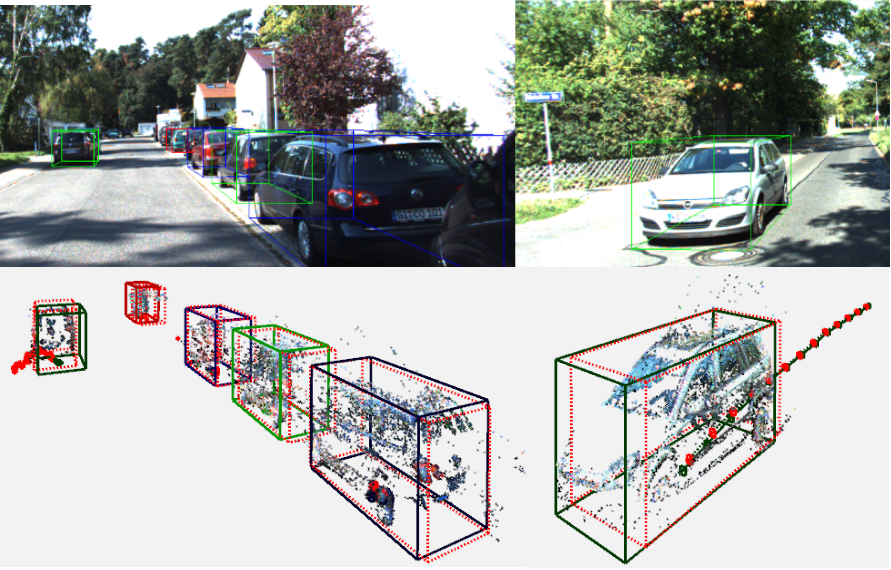}
    \caption{\textbf{Qualitative result of our method.} Top: 3D object proposals are projected and visualized in the image domain. Bottom: We present object detections and accumulated sparse pointclouds in the world coordinate system. Red dashed bounding boxes represent ground truth objects, whereas the red marks on the ground plane indicate ground truth object positions. Object trajectories are globally consistent. For the moving car on the right the track also exhibits local smoothness. Please note that DirectTracker does not estimate camera poses. Thus we utilize estimates from the visual odometry system \cite{engel2017direct} for visualization in the world coordinate system. }
    \label{fig:main_figure}
\end{figure}
To this end, we propose \textbf{DirectTracker}, a simple, yet effective framework that utilizes building blocks of a direct \ac{vo} pipeline \cite{engel2017direct} for the task of 3D object detection and tracking. It is worth noting that we do not address the visual odometry task and do not estimate camera poses with respect to the world coordinate system. Instead, we calculate object's motion in its local coordinate frame (tracking) and localize it (detection) by computing a corresponding 3D transformation with respect to the camera coordinate system. In this formulation we treat every object independently, thus making the whole framework highly parallelizable.

In comparison to existing tracking-by-detection pipelines which utilize 3D proposals from a single-frame 3D object detector \cite{weng2020ab3dmot, weng2020gnn3dmot}, we, firstly, rely only on 2D detections and, secondly, leverage multi-view information to lift localization to 3D and improve 3D tracking. Our sparse reconstructions and final tracks are globally consistent as shown in Fig.~\ref{fig:main_figure}. In addition, due to the refinement with sliding-window bundle adjustment the trajectories of moving cars exhibit local smoothness. Our approach is capable of handling occlusions and perform long-term tracking by effective trajectory management.
\begin{figure*}
    \centering
    \includegraphics[trim={0 1.2cm 0 0},clip,width=0.8\textwidth]{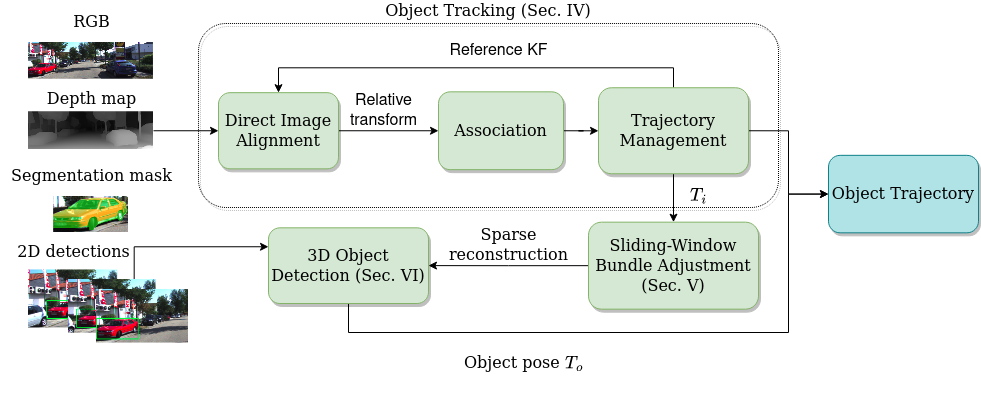}
    \caption{\textbf{The pipeline of our approach.} For simplicity we demonstrate the workflow for a single object. Apart from the association step that is done jointly for all objects, 3D detection and tracking are performed independently and in parallel. In green we denote the core modules, whereas in blue we indicate the output of our method. Please note that in comparison to majority of existing multi-object tracking methods we operate without any 3D object bounding boxes.}
    \label{fig:pipeline}
    \vspace{-0.3cm}
\end{figure*}

The majority of the 3D \ac{mot} benchmarks, including KITTI \cite{geiger2012kitti}, focuses on the 2D evaluation metrics. Nonetheless, as pointed out by \cite{weng2020ab3dmot} the convention of projecting 3D bounding boxes and performing the evaluation in the image space cannot fairly represent the accuracy of tracking in 3D. Instead the authors propose to match detections in 3D and utilize the 3D \ac{iou} metric as a similarity measure. Nevertheless, \ac{iou} does not consider distances between bounding boxes and does not match estimates if their overlap with the ground truth box is zero. This issue is especially apparent for 3D localization of far-away objects, as the depth accuracy decreases nonlinearly with the distance to the stereo camera \cite{pinggera2014limitstereo}. While this measurement error is not of concern for LiDAR-based trackers, stereo-based systems suffer from the introduced bias in the localization. This way even if an object is detected, it can be biased from the ground truth detection due to sensor limitations and, therefore, counted as both a false positive and a false negative, penalizing the performance of the \ac{mot} method. To address this problem we propose to replace 3D \ac{iou} with 3D \ac{giou} \cite{rezatofighi2019giou} as a similarity measure and use \ac{hota} with 3D \ac{giou} as the main metric for the 3D \ac{mot} task. 

In summary, we make the following contributions: 1)~we present a new online\footnote{We adhere to the term \textit{online} as defined in \cite{sharma2018beyond, luo2021motreview}, where online trackers denote methods that utilize data available up-to the current time instant. Please note that the term does not refer to the operating speed, in particular, \textit{online} does not strictly imply real-time execution.} \ac{mot} approach that effectively detects and tracks cars in 3D; 2)~we propose to change the evaluation protocol for the 3D \ac{mot} task based on the recently introduced \ac{hota} metric by integrating 3D \ac{giou} as its similarity measure, which allows to fairly assess the performance of stereo-based tracking methods; 3)~we perform an extensive evaluation of 3D tracking and show competitive results on the KITTI Tracking benchmark \cite{geiger2012kitti}.
\section{Related Work}
\subsection{Vision-based Multi-Object Tracking}
The majority of the existing vision-based multi-object tracking approaches follow the tracking-by-detection paradigm \cite{osep2017ciwt, luiten2020motsfusion, weng2020gnn3dmot, weng2020ab3dmot}, where, firstly, a detector is applied in each frame and, secondly, its estimates are matched into tracks corresponding to a single object. In \cite{osep2017ciwt} an over-complete set of hypotheses for object motion is created based on the 2D detections and 3D object proposals, where the best model is selected based on the 2D-3D Kalman filter. In our work we utilize segmentation masks directly to prevent spurious points from interfering into the 3D object tracking and detection. Another relevant work \cite{luiten2020motsfusion} tracks objects in the image space by warping the corresponding segmentation masks using sparse optical flow. The association is then performed using the Hungarian algorithm \cite{kuhn1955hungarian}. In our work we directly track in 3D and estimate a 6 \ac{dof} rigid transformation, which avoids potential errors caused by inaccurate optical flow. Several joint detection-tracking algorithms have been recently proposed \cite{tokmakov2021permatrack, hu2021quasidense}, which show state-of-the-art \ac{mot} performance. Nevertheless, it is still unclear if they are able generalize to new or under-represented object classes.
\subsection{Objects in Visual Odometry and SLAM}
In contrast to \ac{mot} methods, object-based \ac{slam} intends to jointly estimate camera poses and object landmarks, where the latter are commonly represented as cuboids \cite{salas2013slam++, yang2019cubeslam}, quadrics \cite{nicholson2018quadricslam} or point clusters \cite{huang2019clusterslam}. As the majority of these methods perform evaluation using custom benchmarks, it is difficult to compare their performance against existing \ac{mot} approaches. On the other hand, related approaches in dynamic \ac{slam} \cite{bescos2018dynaslam, ballester2021dot} eliminate information from the moving objects and evaluate system performance mainly against odometry baselines. In our work we leverage several methods from \ac{vo} and \ac{slam} frameworks to continuously track dynamic objects. We further adopt KITTI Tracking benchmark \cite{geiger2012kitti}, which has been widely used by the community for the evaluation of tracking methods.
\input{method}
\section{Experiments}
\begin{table*}[t!]
\centering
\begin{tabular}{|c|c|c|c|c|c|c|c|c|c|}
     \hline
	 \multicolumn{2}{|c|}{Method} & \multirow{2}{*}{HOTA $\uparrow$} & \multirow{2}{*}{DetA $\uparrow$} & \multirow{2}{*}{AssA $\uparrow$} & \multirow{2}{*}{DetRe $\uparrow$} & \multirow{2}{*}{DetPr $\uparrow$} & \multirow{2}{*}{AssRe $\uparrow$} & \multirow{2}{*}{AssPr $\uparrow$} & \multirow{2}{*}{LocA $\uparrow$} \\
	 \cline{1-2}
    Tracking & 3D Detection & & & & & & & & \\
	\hline\hline
	AB3DMOT \cite{weng2020ab3dmot}
	& DSGN \cite{chen2020dsgn}
	& 53.152
	& 46.036
	& 62.530
	& 49.590
	& 77.335
	& 64.859
	& \textbf{88.234}
	& \textbf{83.634} \\
	\hline
	AB3DMOT \cite{weng2020ab3dmot}
	& Stereo-RCNN \cite{li2019stereo}
	& 56.845                            
	& 49.316
	& 66.435
	& 52.498
	& \textbf{77.76}
	& 69.326
	& 85.416
	&  82.091\\
	\hline
	AB3DMOT \cite{weng2020ab3dmot}
	& Frustum Pointnets \cite{qi2018frustum}
	& 56.766
	& 53.405
	& 61.905
	& 60.308
	& 69.886
	& 66.484
	& 80.532
	& 78.919 \\
	\hline
	\multicolumn{2}{|c|}{Ours}
	& \textbf{60.734}                                                  
	& \textbf{56.283}
	& \textbf{68.573}
	& \textbf{64.547}
	& 67.602
	& \textbf{73.187}
	& 81.843 
	& 77.382\\
	\hline
\end{tabular}
\caption{Evaluation of 3D tracking based on 3D \ac{giou}. Our approach outperforms other methods on the overall \ac{hota} metric. The recall is high since we opportunistically track every object. Nonetheless, it impacts the precision levels and overall localization accuracy. The best results per metric are indicated in bold.}
\label{tab:3D_mot_giou}
\vspace{-0.3cm}
\end{table*}
\begin{table*}[t!]
\centering
\begin{tabular}{|c|c|c|c|c|c|c|c|c|c|}
     \hline
	 \multicolumn{2}{|c|}{Method} & \multirow{2}{*}{HOTA $\uparrow$} & \multirow{2}{*}{DetA $\uparrow$} & \multirow{2}{*}{AssA $\uparrow$} & \multirow{2}{*}{DetRe $\uparrow$} & \multirow{2}{*}{DetPr $\uparrow$} & \multirow{2}{*}{AssRe $\uparrow$} & \multirow{2}{*}{AssPr $\uparrow$} & \multirow{2}{*}{LocA $\uparrow$} \\
	 \cline{1-2}
    Tracking & 3D Detection & & & & & & & & \\
	\hline\hline
	AB3DMOT \cite{weng2020ab3dmot}
	& DSGN \cite{chen2020dsgn}
	& 45.215                                                        
	& 37.318
	& 56.236
	& 41.467
	& \textbf{64.668}
	& 59.586
	& \textbf{78.853}
	& \textbf{76.186} \\
	\hline
	AB3DMOT \cite{weng2020ab3dmot}
	& Stereo-RCNN \cite{li2019stereo}
	& \textbf{46.297}                                                                    
	& 38.222
	& 57.322
	& 42.493
	& 62.940
	& 61.657 
	& 74.372
	& 74.504\\
	\hline
	AB3DMOT \cite{weng2020ab3dmot}
	& Frustum Pointnets \cite{qi2018frustum}
	& 44.581                                                         
	& \textbf{38.951}
	& 52.676
	& 47.053 
	& 54.525
	& 58.976
	& 68.736
	& 72.468 \\
	\hline
	\multicolumn{2}{|c|}{Ours}
	& 45.560                                         
	& 37.937
	& \textbf{57.420}
	& \textbf{48.295}
	& 50.574
	& \textbf{63.864}
	& 69.597
	& 72.010\\
	\hline
\end{tabular}
\caption{Evaluation of 3D tracking based on 3D \ac{iou}. We demonstrate how the change of similarity measure affects the performance of our approach as many biased, non-overlapping bounding boxes penalize the accuracy of our tracker. The best results per metric are indicated in bold.}
\label{tab:3D_mot_iou}
\vspace{-0.3cm}
\end{table*}

\begin{table*}[t!]
\centering
\begin{tabular}{|c|c|c|c|c||c|c|c|c|c|}
     \hline
    \multicolumn{2}{|c|}{Method} & \multirow{2}{*}{HOTA $\uparrow$} & \multirow{2}{*}{DetA $\uparrow$} & \multirow{2}{*}{AssA $\uparrow$} & \multirow{2}{*}{MOTA $\uparrow$} & \multirow{2}{*}{MOTP $\uparrow$} & \multirow{2}{*}{IDSW $\downarrow$} & \multirow{2}{*}{FN $\downarrow$} & \multirow{2}{*}{FP $\downarrow$}\\
	 \cline{1-2}
    Tracking & 2D Detection & & & & & & & & \\
	\hline\hline
	CIWT \cite{osep2017ciwt}
	& Track R-CNN \cite{voigtlaender2019trackrcnn}
	& 66.085
	& 62.706
	& 70.798
	& 72.790     
	& 80.222
	& 102
	& 956
	& 989 \\
	\hline
	AB3DMOT \cite{weng2020ab3dmot}
	& DSGN \cite{chen2020dsgn}
	& 54.608        
	& 48.192
	& 62.446
	& 53.037
	& 84.311
	& 150
	& 3041
	& 342 \\
	\hline
	AB3DMOT \cite{weng2020ab3dmot}
	& Stereo-RCNN \cite{li2019stereo}
	& 61.729        
	& 55.796
	& 68.605
	& 64.363    
	& 85.425
	& \textbf{63}
	& 2531
	& 87 \\
	\hline
	AB3DMOT \cite{weng2020ab3dmot}
	& RRC \cite{ren2017rrc} + F-Pointnets \cite{qi2018frustum}
	& 71.671        
	& 76.141
	& 67.607
	& 81.071
	& \textbf{90.879}
	& 97
	& \textbf{121}
	& 151 \\
	\hline
	Ours
	& RRC \cite{ren2017rrc} + BB2SegNet \cite{luiten2018premvos}
	& \textbf{80.913}       
	& \textbf{85.569}
	& \textbf{76.664}
	& \textbf{93.234}
	& 90.500
	& 77
	& 387
	& \textbf{45} \\
	\hline
	\hline
	MOTSFusion \cite{luiten2020motsfusion} (offline)
	& RRC \cite{ren2017rrc} + BB2SegNet \cite{luiten2018premvos}
	& 83.125        
	& 85.797
	& 80.654
	& 94.045
	& 90.612
	& 33
	& 355
	& 33 \\
	\hline
\end{tabular}
\caption{Evaluation of 2D tracking on \ac{hota} and CLEARMOT metrics based on 2D \ac{iou}. Our approach is competitive against other methods across all metrics and demonstrates the performance comparable to the offline MOTSFusion \cite{luiten2020motsfusion}. The best results are indicated in bold.}
\label{tab:2D_mot}
\vspace{-0.3cm}
\end{table*}
\subsection{Dataset}
We choose the KITTI Tracking benchmark \cite{geiger2012kitti} for evaluation of our approach, since, to the best of our knowledge, it is one of the very few existing benchmarks that contains both the stereo images and ground truth object labels. The dataset consists of 21 sequences with, which we split the data into training and validation sets according to \cite{luiten2020motsfusion} to fairly compare against other methods. Specifically, we search for optimal parameter configurations using training sequences and evaluate the performance of our method on the validation set. As our approach follows the principles of rigid body motion, we conduct the evaluation on the Car class objects. Stereo image pairs are required for depth inference by DispNet3 \cite{ilg2018dispnet3}, whereas the main pipeline operates on the monocular sequences.
\subsection{System Setup and Evaluated Methods}
As an input to our pipeline we follow MOTSFusion \cite{luiten2020motsfusion} setup. We utilize 2D detections from RRC \cite{ren2017rrc} and the corresponding segmentation masks provided by \cite{luiten2020motsfusion}. Stereo-based depth maps are obtained from DispNet3 \cite{ilg2018dispnet3}. As for hyperparameters, we keep Age$_\text{max} = 2$, $w_1 = 5$, $w_2 = 3$, $w_3 = 1$ and $\lambda = 1.5$, which empirically achieve the best performance. The minimal similarity threshold for the association based on 2D \ac{iou} is set to $0.05$. The number of keyframes in the sliding window is restricted to $6$. For the keyframe generation we keep the threshold of $0.5$.

In the evaluation against other methods we select two stereo-based 3D object detectors, namely Stereo R-CNN and DSGN, which we re-train on our training split based on the default configurations provided by the authors. As a pointcloud-based detector, we use Frustum Pointnets \cite{qi2018frustum} trained on the chosen training split using the stereo-depth from DispNet3 \cite{ilg2018dispnet3}. All detections are further fed into the tracking method, namely AB3DMOT \cite{weng2020ab3dmot}, and the final results are evaluated. We preprocess the DSGN output and filter the detections by confidence score before tracking. The threshold is chosen to maximize the final 3D \ac{mot} metric. For the 2D \ac{mot} task we include the CIWT \cite{osep2017ciwt} framework.

\subsection{Evaluation Metrics}
We adopt the latest protocol standards for analyzing 3D \ac{mot} methods from the KITTI Tracking benchmark \cite{geiger2012kitti} and utilize \ac{hota} \cite{luiten2021hota} metrics along with the similarity (alpha) thresholds provided by the authors. We further demonstrate the effect of changing the similarity measure from \ac{iou} to \ac{giou} on the 3D performance of vision-based trackers. We normalize \ac{giou} values from the range $[-1, 1]$ to $[0, 1]$ to comply with the \ac{hota} similarity thresholds. For 2D \ac{mot} we additionally report the evaluation on the CLEARMOT \cite{bernardin2008clearmot} metrics such as \ac{mota}, \ac{motp}, number of \ac{idsw}, \ac{fp} and \ac{fn}. We use the TrackEval evaluation tool \cite{luiten2020trackeval} and extend it with the 3D \ac{giou} computations.
\begin{table*}
\centering
\begin{tabular}{|c|c|c|c|c|c|c|}
    \hline
	Configuration & HOTA(3D) $\uparrow$ & DetA(3D) $\uparrow$ & AssA(3D) $\uparrow$ & HOTA(2D) $\uparrow$ & DetA(2D) $\uparrow$ & AssA(2D) $\uparrow$\\
	\hline\hline
	mono-depth (Adabins \cite{bhat2021adabins})
	& 50.092                          
	& 41.977
	& 63.860
	& 80.670
	& 85.588
	& 76.190\\
	\hline
    grayscale \ac{dia}
	& 60.585                                  
	& 55.972
	& 68.522
	& 80.683        
	& 85.528
	& 76.273\\
	\hline
	no 2D tracking
	& 60.225                                     
	& 56.454
	& 67.197
	& 78.020         
	& 84.477
	& 72.229 \\
	\hline
	no \ac{ba} refinement (\ref{sec:bundle_adjustment})
	& 55.405                      
	& 51.390
	& 62.902
	& 80.802        
	& 85.566
	& 76.460\\
	\hline
    no 3D box optimization (\ref{subsec:optim_bbox})
	& 51.763                                    
	& 46.180
	& 60.562
	& 67.496        
	& 67.842
	& 68.378\\
	\hline
    no 3D-2D (\ref{subsubsec:3d2d})
	& 57.119                                  
	& 51.014 
	& 66.947
	& 80.678        
	& 85.617
	& 76.179\\
	\hline
    no 3D-3D (\ref{subsubsec:3d3d})
	& 52.255                                   
	& 44.035
	& 66.623
	& 80.937        
	& 85.658
	& 76.631\\
	\hline
	no regularisation (\ref{subsubsec:reg})
	& 59.682 
	& 55.773    
	& 66.548    
	& 80.869    
	& 85.554    
	& 76.595\\
	\hline
    no F-Pointnets \cite{qi2018frustum} 3D box dims
	& 59.877                           
	& 55.994
	& 66.892
	& 80.857        
	& 85.546
	& 76.581\\
	\hline
    no fusion (\ref{subsec:fusion_bbox})
	& 56.151                                     
	& 52.814
	& 61.263
	& 80.666        
	& 85.543
	& 76.222\\
	\hline
	\hline
	full system
	& 60.734                                                 
	& 56.283
	& 68.573
	& 80.913
	& 85.569
	& 76.664\\
	\hline
\end{tabular}
\caption{Ablation study. We present the performance of our method with different components of the pipeline removed or replaced. The evaluation is conducted for both 3D (\ac{giou}) and 2D tracking (\ac{iou}).}
\vspace{-0.5cm}
\label{tab:ablstudy}
\end{table*}
\subsection{3D Multi-Object Tracking}
In Table \ref{tab:3D_mot_giou} we demonstrate our evaluation results on the 3D \ac{mot} task. Our approach significantly outperforms other methods on HOTA and its decomposites, namely the detection accuracy (DetA) and the association accuracy (AssA). We consistently report detections even for the difficult cases, such as far-away objects, therefore the recall values for detection and association remain high. At the same time it affects the precision values and localization accuracy (LocA) with biased detections. Estimating the uncertainty of object detections and pruning them based on a fixed confidence threshold would alleviate the issue in compromise of achieving lower recall values.

In Table \ref{tab:3D_mot_iou} we also provide the results using 3D \ac{iou} as a similarity metric. Our method is affected the most with the $\sim$25\% decrease in \ac{hota} value, as many biased object proposals are not matched and negatively contribute as false positives and false negatives towards the metric value. A similar impact can be observed from the Frustum-Pointnets \cite{qi2018frustum} baseline ($\sim$21 \%), since it is trained on the same depth maps that are utilized in our approach. The impact is not as severe for DSGN ($\sim$14 \%) and Stereo R-CNN ($\sim$18 \%) baselines.

Although all methods are affected to different extent, it is clear that the baselines with higher recall are penalized greater than the ones that assign low confidence to inaccurate detections and prune them before tracking. This observation is also reported by \cite{pang2021simpletrack} using different benchmarks and CLEARMOT metrics. We are convinced that in order to obtain an online tracker that has both high recall and high precision we need to change the way we evaluate its performance. Replacing \ac{iou} by \ac{giou}, as we propose, and using \ac{hota} as an evaluation metric gives a better estimate of the overall tracking performance.

\subsection{2D Multi-Object Tracking}
In this section we provide the results for the 2D \ac{mot} task. As shown in Table \ref{tab:2D_mot} our method outperforms other approaches over all HOTA metrics. For the MOTA metrics we show competitive performance, which is consistently ranked first or second among the evaluated approaches. It should be noted that although our method operates online, its performance is comparable to the offline\footnote{In accordance with the definition of \textit{online} methods we refer to \textit{offline} approaches as those that operate on data collected from all frames in the sequence \cite{luo2021motreview}.} MOTSFusion \cite{luiten2020motsfusion}. 

\subsection{Ablation Study}
In this section we analyze the effect of different parts of our pipeline on the final performance. The results are summarized in Table \ref{tab:ablstudy}. In each row we report the performance of our method with the changes given in the first column. In ``mono-depth" version of our approach we replace the stereo-depth from DispNet3 \cite{ilg2018dispnet3} by monocular estimates from Adabins \cite{bhat2021adabins}. For ``grayscale \ac{dia}" we perform alignment based on 1-channel instead of 3-channels images. Other configurations specify a component that is removed from the pipeline.

As it can be observed from Table \ref{tab:ablstudy} the depth quality has a significant impact on the performance of our approach in 3D, since we heavily rely on 3D cues for the object detection. In ``no 2D tracking" we experiment with our method by terminating all tracks, where object is not successfully tracked in 3D with direct image alignment. As anticipated, it degrades the tracking performance and penalizes the accuracy with false negatives. The bounding box refinement as proposed in Sec. \ref{subsec:optim_bbox} is shown to add significant boost to the object detection module using both 3D and 2D metrics. It is worth noting that the combination of optimization terms based on the 3D and 2D cues demonstrate its complementary effect on the object localization.

As the impact of switching from the color-based to the grayscale \ac{dia} is not significant, we show that our approach can work with the grayscale video streams at a similar accuracy level. Although the final version of our approach utilizes the bounding box dimensions obtained from the Frustum-Pointnets\cite{qi2018frustum} detections, we demonstrate that our method is capable of delivering competitive results even without them. 

\section{Conclusion}
In this paper we have presented DirectTracker, an online \ac{mot} approach that utilizes direct image alignment and photometric bundle adjustment for the task of 3D object detection and tracking. We propose to replace restrictive \ac{iou} as a similarity measure by \ac{giou}, which allows to fairly evaluate the performance of stereo-based trackers. Through our experiments we show competitive performance of our method on the KITTI benchmark in comparison to other tracking-by-detection approaches. Our ablation study demonstrates the effect of every component that contributes to the final tracking accuracy in 3D and 2D. We hope that our method has revealed the power of direct methods utilized in the scope of multi-object tracking and that it will promote further research in this direction. In particular, evaluation of our method on a different dataset (e.g. Argoverse 2 \cite{Argoverse2}) is an interesting exploratory direction, which remains part of the future work. Moreover, our approach can be extended to support object's motion in 6 \ac{dof} by incorporating a more sophisticated initialization method for 3D object detection (e.g. the QuickHull algorithm \cite{barber1996quickhull}), which does not rely on projection of points onto the ground plane and directly works in 3D, and including roll and pitch parameters into the optimization stage. A suitable benchmark would adequately complement such extension.

\bibliographystyle{ieeetr}
\bibliography{references}
\end{document}

%% file: method.tex
\section{System Overview}
In the following sections we will give a detailed overview of the proposed system as demonstrated in Fig. \ref{fig:pipeline}. In particular, we will describe the three major modules of our pipeline: 1) the object tracking module that performs frame-to-frame 3D object tracking and association (Sec. \ref{sec:object_tracking}); 2) the object-based \ac{ba} module that integrates relative odometry constraints from the tracking module and jointly refines object poses and 3D reconstruction in a local sliding window (Sec. \ref{sec:bundle_adjustment}); 3) the 3D object detection module which leverages tracking and reconstruction estimates to obtain globally consistent object poses (Sec. \ref{sec:object_detection}). Object trajectories are estimated in parallel and independently from each other, except for the association step that is performed jointly for all 2D detections. In this work we do not tackle the visual odometry task, therefore all object poses are estimated in the camera coordinate system.

As an input to our tracking module we utilize a video image sequence, depth maps and semantic segmentation masks. The latter can be obtained from any off-the-shelf segmentation network, since our method does not require pixel-accurate masks to reliable track objects. In most of the experiments we use stereo depth (e.g. predicted by DispNet3 \cite{ilg2018dispnet3}), however our pipeline is able to work with the depth from different sources such as single camera, LiDAR and \ac{sfm}. To obtain amodal 3D bounding boxes we additionally supply 2D detection estimates to the 3D object detection module. If external 3D object detections are available, they serve as a prior in the estimation of accurate object dimensions.
\section{Object Tracking}
\label{sec:object_tracking}
Object tracking is carried out in a two-step manner. Firstly, 3D tracking is attempted by two-frame \ac{dia} in a coarse-to-fine optimization scheme (Sec. \ref{subsec:dia}).
Secondly, the association is performed in the image space using the Hungarian algorithm \cite{kuhn1955hungarian} based on the \ac{iou} score between the warped and the input segmentation masks (Sec. \ref{subsec:ass}).
\subsection{Direct Image Alignment}
\label{subsec:dia}
Given a depth image $\mathbf{D}_t$ and a segmentation mask $\Omega_t$ we seek to find a rigid body transformation $\mathbf{T}_t^{t - 1} \in SE(3)$ that minimizes the photometric error: 
\begin{equation}
\label{eq:dia}
\arg\!\min_{\mathbf{T}_t^{t - 1} \in SE(3)} \sum_{\mathbf{p} \in \Omega_t }|| \mathbf{I}_{t - 1}(\mathbf{p}') - \mathbf{I}_{t}(\mathbf{p}) ||_{\gamma }\, ,
\end{equation}
where $\mathbf{p}$ is a pixel which lies within a segmentation mask $\Omega_t$, $\mathbf{I}_{t - 1}$ and $\mathbf{I}_t$ are the previous and the current image frames, respectively, $||\cdot||_{\gamma}$ is a Huber norm. The warped pixel coordinates $\mathbf{p'}$ can be then defined as:
\begin{equation}
    \label{eq:warped_pixel}
    \mathbf{p'} = \pi(\mathbf{T}^{t - 1}_t\pi^{-1}(\mathbf{p}; \mathbf{D}_t(\mathbf{p}))) \, ,
\end{equation}
with the perspective projection function $\pi$. To increase the radius of the convergence basin the optimization is undertaken in a coarse-to-fine manner. Since distant objects occupy little image space, the conventional pyramid construction with the image downsampling as done in \cite{engel2014lsd, engel2017direct} might significantly affect the alignment success and degrade the pose accuracy. Therefore, we adapt the scaling based on the minimal object size to guarantee that relevant information is not over-smoothed on the coarse levels. The pose is iteratively refined at each image pyramid level using the Levenberg-Marquardt optimization algorithm \cite{levenberg1944levenbergmarquardt}. In case the optimization fails or too few pixels correspond to an object, we perform 2D tracking using sparse optical flow, as it is done in \cite{luiten2020motsfusion}. If 2D tracking is unsuccessful as well, the object is considered to be lost and its potential future re-appearance would initiate a new track. 

\ac{dia} requires good relative pose initialization. For new tracks a set of pose candidates is suggested and the winner is chosen based on the final cost value. For the sake of efficiency this selection is executed only on the sparsest level of the image pyramid. In case the object has been successfully tracked in 3D before, a constant motion model based on the rigid transformation $\mathbf{T}^{t-2}_{t-1}$ is utilized instead.
\subsection{2D Association and Trajectory Management}
\label{subsec:ass}
For the association step we leverage estimated relative object motion and warp the segmentation mask to the previous frame at time $t - 1$. The Hungarian algorithm \cite{kuhn1955hungarian} is utilized and one-to-one matches are obtained based on the 2D \ac{iou} metric. A correspondence is accepted, if the similarity score is above a minimal threshold. Successfully matched masks from the current frame receive a track id of their match and their tracklet is merged into the corresponding trajectory. In case an object does not receive a match, we see whether it has been tracked successfully in 3D before. If it is so, the object is pertained in memory with the corresponding warped mask. This allows to handle the cases, when the input segmentation masks are not temporarily consistent. If an object has never been tracked before, a new trajectory is initiated. Similarly to AB3DMOT \cite{weng2020ab3dmot} the trajectory is terminated if it has not been extended by any new detections for $\text{Age}_\text{max}$ frames.
\section{Sliding-Window Bundle Adjustment}
\label{sec:bundle_adjustment}
If an object is successfully tracked in 3D and the magnitude of its apparent motion is above a minimal threshold, we can promote the current frame to be a keyframe. Although we only consider car objects, the heuristic can be extended to the multi-class tracking by setting a class-specific threshold. Inspired by DSO \cite{engel2017direct}, a state-of-the-art direct visual odometry algorithm, we jointly refine the object poses and 3D points in a keyframe-based windowed \ac{ba}. Similarly to \cite{engel2017direct} a sparse set of points is selected uniformly across the whole region defined by the object's segmentation mask. The photometric energy function can be formulated as
\begin{align}
    \label{eq:ba}
    \text{E}_\text{photo} &= \sum_{i \in \mathcal{F}} \sum_{\mathbf{q} \in \mathcal{P}_i} \sum_{j \in \mathcal{Q}} \text{E}_{i,\mathbf{q},j} \, , \text {    where}\\
    \text{E}_{i,\mathbf{q},j} = \sum_{\mathbf{p} \in \mathcal{N}_\mathbf{q}} &\Bigg \vert \Bigg \vert  I_j(\mathbf{p'}) - \frac{\sum_{\mathbf{p} \in \mathcal{N}_\mathbf{q}} I_j(\mathbf{p'})}{ \sum_{\mathbf{p} \in \mathcal{N}_\mathbf{q}} I_i(\mathbf{p})} I_i(\mathbf{p}) \Bigg \vert \Bigg \vert_{\gamma} \, ,
\end{align}
with $\mathcal{F}$ defined as the set of keyframes in the object's trajectory, $\mathcal{P}_i$ as the set of 2D points in the keyframe $i$ that correspond to the object, $\mathcal{Q}$ as the set of keyframes in the object's trajectory that co-observe a point $\mathbf{q}$ and $||\cdot||_{\gamma}$ as the Huber norm. The point $\mathbf{p'}$ is defined analogous to \eqref{eq:warped_pixel}, where the relative transformation $\mathbf{T}_{i,j} := \mathbf{T}_j^{-1}\mathbf{T}_i$ from the coordinate frame of a keyframe $i$ to the coordinate frame of a keyframe $j$ is considered instead. The poses $\mathbf{T}_{i,j} $ are defined with respect to a common local coordinate frame in the origin of the object's trajectory and initialized by concatenating frame-to-frame estimates from the \ac{dia}. For energy computation we utilize a patch of neighboring pixels $\mathcal{N}_\mathbf{q}$ based on the residual pattern proposed in \cite{engel2017direct} that has been shown to be a good trade-off between robustness to motion blur, computational efficiency and amount of contained information. Since the affine brightness parameters are not estimated, we leverage the \ac{lssd} function \cite{roma2002comparative} to overcome the potential brightness and contrast changes. To keep the size of the sliding window bounded we discard the oldest keyframes from the optimization. One can alternatively introduce a marginalization prior as in \cite{engel2017direct} and keep the constraints between active and removed variables as an additional energy term, which remains part of the future work.
\section{3D Object Detection}
\label{sec:object_detection}
During \ac{dia} and sliding-window bundle adjustment object poses are defined in a local coordinate system with respect to the origin of its trajectory, i.e. in the reference frame. In order to obtain object poses in the camera coordinate system we need to compute a transformation between camera and the object's local coordinate system. This task is achieved by the object detection module that estimates the aforementioned transformation at a keyframe rate. Moreover, we restrict the transformation to be a 4 \ac{dof} homogeneous matrix by estimating a 3D translational vector and a yaw angle. This simplified parameterization has shown to be sufficient in our evaluations on the KITTI dataset \cite{geiger2012kitti}, where the ground plane is assumed to be flat and camera pitch is considered to be negligible. We model the objects as bounding boxes (cuboids) and additionally estimate the corresponding dimensions. 

Object detection is carried out as a three-step procedure. Firstly, an object's bounding box is regressed in the bird's-eye view given sparse active 3D points from the windowed \ac{ba} (Sec. \ref{subsec:convex_hull}). Secondly, the detection estimate is refined by leveraging complementary 2D and 3D cues in a joint optimization cost function (Sec. \ref{subsec:optim_bbox}). Thirdly, we exploit the fact that the object's transformation from the camera coordinate system is computed in every keyframe by propagating new estimates to the past keyframes and fusing them based on their uncertainty values (Sec. \ref{subsec:fusion_bbox}).

\subsection{Convex-Hull Bounding Box Regression}
\label{subsec:convex_hull}
As a method for initialization of a 3D object proposal we opt for a simple and fast regression algorithm based on the principles of occupancy grids and convex-hull estimation. Firstly, we accumulate the pointcloud in the reference frame, select a set of visible and active 3D points from the sliding-window \ac{ba} and project them to the ground plane. The projected points are further allocated into a 2D occupancy grid. A grid cell is considered ``occupied'' if the number of residing points exceeds a minimal count. Based on the occupancy information a convex hull is constructed and a number of enclosing rotated 2D bounding boxes are proposed. The bounding box is parameterized by an angle $\theta$, 2D centre location and 2D dimensions (width and length). The optimal solution minimizes the sum of squared distances between each point in the 2D grid and the bounding box boundary. An estimated 2D box is lifted to 3D by considering the median y-coordinate of all 3D points and the car-class average height value. In case external 3D detections are available, we directly consider corresponding 3D dimensions before passing 3D bounding box for optimization.
\subsection{Optimization-based Bounding Box Refinement}
\label{subsec:optim_bbox}
Inspired by CubeSLAM \cite{yang2019cubeslam} we introduce a nonlinear least squares optimization function, which integrates multi-view 2D and 3D information to refine the 3D object proposal from the previous step. To this end, we utilize amodal 2D bounding boxes, object odometry poses and sparse pointcloud obtained from the sliding-window bundle adjustment to effectively constrain the 3D location and orientation angle of an object in the camera coordinate system. As we deal with traffic scenarios, objects are rarely observed from all viewing directions and the bounding box dimensions are not well-constrained in most of cases unless a shape prior is provided \cite{sharma2018beyond, wang2020directshape}. Therefore, we do not optimize over the dimensional variables and keep them fixed during the optimization.

Since we assume that both the camera and the observed object move in 3D and have a single rotational axis, we estimate $\mathbf{T}_o$ as a 4 \ac{dof} matrix. The variable represents the transformation from the object's local coordinate system in the reference frame to the camera coordinate system. The objective is formulated as 
\begin{equation}
    \label{eq:optim_bbox3d}
    \arg\!\min_{\mathbf{T}_o} \{w_1\text{E}_\text{3D-2D} + W(\text{E}_\text{3D-3D})\text{E}_\text{3D-3D} + w_3\text{E}_\text{reg} \} \, ,
\end{equation}
where $W(\text{E}_\text{3D-3D})$ is a dynamic weight that can be defined according to \cite{stumberg2022dmvio} as
\begin{equation}
    \label{eq:dynamic_weight}
    W(\text{E}_\text{3D-3D}) = w_2 \cdot \begin{cases} 
      (\lambda^2 \cdot n) / \text{E}_\text{3D-3D}, & \text{if  } |\text{E}_\text{3D-3D}| \geq \lambda^2 \cdot n \\
      1, & \text{otherwise}\,.
   \end{cases}
\end{equation}
In Eq.~\eqref{eq:dynamic_weight} we define $n$ as the number of the 3D-3D residuals, $\lambda$ and $w_2$ as static weights. By introducing the dynamic weight we address the cases when the 3D reconstruction is poor and, therefore, increase the relative contribution from the other terms. The optimization is performed using the Levenberg-Marquardt algorithm \cite{levenberg1944levenbergmarquardt}. In the following sections we describe in detail each of the optimization terms.
\subsubsection{3D-3D term}
\label{subsubsec:3d3d}
Similarly to \cite{yang2019cubeslam} we introduce the constraint between an object and its 3D reconstruction. In particular, the 3D points $\mathcal{P}^{3D}$ which belong to the object must lie within its bounding box upon transforming them to the object's local coordinate system, as formulated in Eq.~\eqref{eq:optim_bbox3d_3dterm}, where  $\mathcal{F}$ is a set of keyframes, $\mathbf{T}_i$ is the transformation from the keyframe $i$ to the reference frame obtained from \ac{ba} and $\mathbf{d} \in {\rm I\!R^3}$ denotes object's 3D dimensions, i.e. its width, height and length.
\begin{equation}
    \label{eq:optim_bbox3d_3dterm}
    \text{E}_\text{3D-3D} := \sum_{i \in \mathcal{F}} \sum_{\mathbf{x} \in \mathcal{P}_i^\text{3D}} \Bigg \vert \Bigg \vert \max \left\{ \mathbf{0}, \frac{|\mathbf{T}^{-1}_o \mathbf{T}_i \mathbf{x}| - \frac{1}{2}\mathbf{d}}{\sigma_\mathbf{x}} \right\} \Bigg \vert \Bigg \vert_\gamma \, .
\end{equation}
The term is complementary to the previous step (Sec. \ref{subsec:convex_hull}), where the 3D points projected to the ground plane are considered for the bounding box regression. In addition, we include the depth uncertainty $\sigma_\mathbf{x}$ obtained from the sliding-window \ac{ba}. This weight helps to reduce the impact of points with high photometric error \cite{engel2013semi}.
\subsubsection{3D-2D term}
\label{subsubsec:3d2d}
Segmentation masks provide only visible information about the 3D object. Therefore, they are not sufficient for the estimation of the full-size 3D bounding boxes. To this end, we utilize amodal 2D bounding boxes, which are easy to obtain by deploying existing 2D object detectors. We project the estimated 3D bounding box to the image plane and evaluate how well it agrees with the 2D detection. Moreover, since we consistently track objects, we can leverage 2D detections from multiple views along the trajectory. In most of the cases, this strategy gives better constraining power to the object's 3D position. Analogously to the camera-2D measurement term in \cite{yang2019cubeslam} we define a multi-view 3D-2D optimization term as
\begin{equation}
    \label{eq:optim_bbox3d_2dterm}
    \text{E}_\text{3D-2D} := \sum_{i \in \mathcal{B}} \vert \vert \phi(\{\pi(\mathbf{T}^{-1}_i\mathbf{T}_o \mathbf{c}_k)\}_{k \in [1 .. 8]}) - \mathbf{b}_i \vert \vert \, ,
\end{equation}
where $\mathcal{B}$ is a set of keyframes that have an amodal 2D bounding box detection $\mathbf{b}$, $\mathbf{c} \in [ \pm d_x / 2, \pm d_y / 2, \pm d_z / 2]^T$ corresponds to 8 corners of a 3D bounding box with dimensions $\mathbf{d}$ in the origin of the object's local coordinate system, $\phi(\cdot)$ is a function that extracts enclosing axis-aligned 2D bounding box parameters (center and dimensions) around the projected corners. Although Eq. \eqref{eq:optim_bbox3d_2dterm} has many valid solutions, it imposes a strong constraint on the object's relative position with respect to the camera, which allows to estimate an amodal 3D bounding box even when the 3D data is incomplete due to occlusions.
\subsubsection{Regularization term}
\label{subsubsec:reg}
Since 3D object detection is performed at keyframe rate, we can leverage the last-estimated object proposal and impose its pose $\mathbf{T}'_o$ as a regularization during the optimization. Specifically, the corresponding term can be expressed as
\begin{equation}
    \label{eq:prior}
    \text{E}_\text{reg} := ||\text{Log}(\mathbf{T}^{-1}_o\mathbf{T'}_o)|| \, ,
\end{equation}
where Log$(\cdot)$ is a mapping from an element of the Lie group in SE(3) to its twist coordinates in $\mathfrak{se}(3)$.
\subsection{3D Object Proposal Fusion}
\label{subsec:fusion_bbox}
Far-away objects pose difficulties for our optimization-based detector as the error of stereo-depth grows quickly with the distance \cite{pinggera2014limitstereo}. In addition, distant objects occupy little image space, thus their 3D reconstruction is very sparse. Nonetheless, as they approach the ego-vehicle, we accumulate more accurate 3D points and increase the confidence of our object proposal. Since 3D object detection is performed every keyframe, we can integrate new proposals into the past estimates based on their uncertainties. To this end, we compute the weighted average between the current estimate (measurement) and all the past estimates (prior) by utilizing a diagonal covariance matrix $\mathbf{\Sigma}_m$ of the object's location and an error $\sigma_m$ of the orientation angle from the optimization-based refinement step (Sec. \ref{subsec:optim_bbox}). If we assume a Gaussian distribution of errors and omit nonlinear operators such as projection function $\pi$, the fusion corresponds to the Kalman filter update step for uncorrelated variables \cite{rojas2003kalmanfilter}. 
Mathematically we can formulate the update for object's position $\hat{c}$ as
\begin{align}
    \label{eq:fusion}
    \mathbf{\hat{c}} &= (\mathbf{\Sigma}_{p} + \mathbf{\Sigma}_{m})^{-1}(\mathbf{\Sigma}_{p}\mathbf{c}_{m} + \mathbf{\Sigma}_{m}\mathbf{c}_{p}) \, ,\\
    &\mathbf{\hat{\Sigma}} = (\mathbf{I} - \mathbf{\Sigma}_{p} (\mathbf{\Sigma}_{p} + \mathbf{\Sigma}_{m})^{-1})\mathbf{\Sigma}_{p} \, ,
\end{align}
where $\mathbf{c}_{p}$ is a prior position obtained by fusing past estimates, $\mathbf{c}_{m}$ is a new position measurement and $\mathbf{\Sigma}_{p}$ is a prior covariance matrix. 
Similarly the update is carried out for the orientation angle with the additional wrapping function as proposed in \cite{markovic2016wrapping}.

%% file: root.bbl
\begin{thebibliography}{10}

\bibitem{li2019stereo}
P.~Li, X.~Chen, and S.~Shen, ``Stereo r-cnn based 3d object detection for
  autonomous driving,'' in {\em Proceedings of the IEEE/CVF Conference on
  Computer Vision and Pattern Recognition}, pp.~7644--7652, 2019.

\bibitem{qi2018frustum}
C.~R. Qi, W.~Liu, C.~Wu, H.~Su, and L.~J. Guibas, ``Frustum pointnets for 3d
  object detection from rgb-d data,'' in {\em Proceedings of the IEEE
  conference on computer vision and pattern recognition}, pp.~918--927, 2018.

\bibitem{chen2020dsgn}
Y.~Chen, S.~Liu, X.~Shen, and J.~Jia, ``Dsgn: Deep stereo geometry network for
  3d object detection,'' in {\em Proceedings of the IEEE/CVF conference on
  computer vision and pattern recognition}, pp.~12536--12545, 2020.

\bibitem{sun2020disprcnn}
J.~Sun, L.~Chen, Y.~Xie, S.~Zhang, Q.~Jiang, X.~Zhou, and H.~Bao, ``Disp r-cnn:
  Stereo 3d object detection via shape prior guided instance disparity
  estimation,'' in {\em Proceedings of the IEEE/CVF conference on computer
  vision and pattern recognition}, pp.~10548--10557, 2020.

\bibitem{weng2020gnn3dmot}
X.~Weng, Y.~Wang, Y.~Man, and K.~Kitani, ``Gnn3dmot: Graph neural network for
  3d multi-object tracking with multi-feature learning,'' {\em arXiv preprint
  arXiv:2006.07327}, 2020.

\bibitem{weng2020ab3dmot}
X.~Weng, J.~Wang, D.~Held, and K.~Kitani, ``3d multi-object tracking: A
  baseline and new evaluation metrics,'' in {\em Proceedings of the 2020
  IEEE/RSJ International Conference on Intelligent Robots and Systems (IROS)},
  pp.~10359--10366, IEEE, 2020.

\bibitem{kim2021eagermot}
A.~Kim, A.~O{\v{s}}ep, and L.~Leal-Taix{\'e}, ``Eagermot: 3d multi-object
  tracking via sensor fusion,'' in {\em 2021 IEEE International Conference on
  Robotics and Automation (ICRA)}, pp.~11315--11321, IEEE, 2021.

\bibitem{engel2014lsd}
J.~Engel, T.~Sch{\"o}ps, and D.~Cremers, ``Lsd-slam: Large-scale direct
  monocular slam,'' in {\em European conference on computer vision},
  pp.~834--849, Springer, 2014.

\bibitem{engel2017direct}
J.~Engel, V.~Koltun, and D.~Cremers, ``Direct sparse odometry,'' {\em IEEE
  transactions on pattern analysis and machine intelligence}, vol.~40, no.~3,
  pp.~611--625, 2017.

\bibitem{salas2013slam++}
R.~F. Salas-Moreno, R.~A. Newcombe, H.~Strasdat, P.~H. Kelly, and A.~J.
  Davison, ``Slam++: Simultaneous localisation and mapping at the level of
  objects,'' in {\em Proceedings of the IEEE conference on computer vision and
  pattern recognition}, pp.~1352--1359, 2013.

\bibitem{bescos2018dynaslam}
B.~Bescos, J.~M. F{\'a}cil, J.~Civera, and J.~Neira, ``Dynaslam: Tracking,
  mapping, and inpainting in dynamic scenes,'' {\em IEEE Robotics and
  Automation Letters}, vol.~3, no.~4, pp.~4076--4083, 2018.

\bibitem{yang2019cubeslam}
S.~Yang and S.~Scherer, ``Cubeslam: Monocular 3-d object slam,'' {\em IEEE
  Transactions on Robotics}, vol.~35, no.~4, pp.~925--938, 2019.

\bibitem{geiger2012kitti}
A.~Geiger, P.~Lenz, and R.~Urtasun, ``Are we ready for autonomous driving? the
  kitti vision benchmark suite,'' in {\em Proceedings of the 2012 IEEE
  conference on computer vision and pattern recognition}, pp.~3354--3361, IEEE,
  2012.

\bibitem{pinggera2014limitstereo}
P.~Pinggera, D.~Pfeiffer, U.~Franke, and R.~Mester, ``Know your limits:
  Accuracy of long range stereoscopic object measurements in practice,'' in
  {\em European conference on computer vision}, pp.~96--111, Springer, 2014.

\bibitem{rezatofighi2019giou}
H.~Rezatofighi, N.~Tsoi, J.~Gwak, A.~Sadeghian, I.~Reid, and S.~Savarese,
  ``Generalized intersection over union: A metric and a loss for bounding box
  regression,'' in {\em Proceedings of the IEEE/CVF conference on computer
  vision and pattern recognition}, pp.~658--666, 2019.

\bibitem{sharma2018beyond}
S.~Sharma, J.~A. Ansari, J.~K. Murthy, and K.~M. Krishna, ``Beyond pixels:
  Leveraging geometry and shape cues for online multi-object tracking,'' in
  {\em 2018 IEEE International Conference on Robotics and Automation (ICRA)},
  pp.~3508--3515, IEEE, 2018.

\bibitem{luo2021motreview}
W.~Luo, J.~Xing, A.~Milan, X.~Zhang, W.~Liu, and T.-K. Kim, ``Multiple object
  tracking: A literature review,'' {\em Artificial Intelligence}, vol.~293,
  p.~103448, 2021.

\bibitem{osep2017ciwt}
A.~Osep, W.~Mehner, M.~Mathias, and B.~Leibe, ``Combined image-and world-space
  tracking in traffic scenes,'' in {\em Proceedings of the 2017 IEEE
  International Conference on Robotics and Automation (ICRA)}, pp.~1988--1995,
  IEEE, 2017.

\bibitem{luiten2020motsfusion}
J.~Luiten, T.~Fischer, and B.~Leibe, ``Track to reconstruct and reconstruct to
  track,'' {\em IEEE Robotics and Automation Letters}, vol.~5, no.~2,
  pp.~1803--1810, 2020.

\bibitem{kuhn1955hungarian}
H.~W. Kuhn, ``The hungarian method for the assignment problem,'' {\em Naval
  research logistics quarterly}, vol.~2, no.~1-2, pp.~83--97, 1955.

\bibitem{tokmakov2021permatrack}
P.~Tokmakov, J.~Li, W.~Burgard, and A.~Gaidon, ``Learning to track with object
  permanence,'' in {\em Proceedings of the IEEE/CVF International Conference on
  Computer Vision}, pp.~10860--10869, 2021.

\bibitem{hu2021quasidense}
H.-N. Hu, Y.-H. Yang, T.~Fischer, T.~Darrell, F.~Yu, and M.~Sun, ``Monocular
  quasi-dense 3d object tracking,'' {\em arXiv preprint arXiv:2103.07351},
  2021.

\bibitem{nicholson2018quadricslam}
L.~Nicholson, M.~Milford, and N.~S{\"u}nderhauf, ``Quadricslam: Dual quadrics
  from object detections as landmarks in object-oriented slam,'' {\em IEEE
  Robotics and Automation Letters}, vol.~4, no.~1, pp.~1--8, 2018.

\bibitem{huang2019clusterslam}
J.~Huang, S.~Yang, Z.~Zhao, Y.-K. Lai, and S.-M. Hu, ``Clusterslam: A slam
  backend for simultaneous rigid body clustering and motion estimation,'' in
  {\em Proceedings of the IEEE/CVF International Conference on Computer
  Vision}, pp.~5875--5884, 2019.

\bibitem{ballester2021dot}
I.~Ballester, A.~Fontan, J.~Civera, K.~H. Strobl, and R.~Triebel, ``Dot:
  dynamic object tracking for visual slam,'' in {\em Proceedings of the 2021
  IEEE International Conference on Robotics and Automation (ICRA)},
  pp.~11705--11711, IEEE, 2021.

\bibitem{ilg2018dispnet3}
E.~Ilg, T.~Saikia, M.~Keuper, and T.~Brox, ``Occlusions, motion and depth
  boundaries with a generic network for disparity, optical flow or scene flow
  estimation,'' in {\em Proceedings of the European Conference on Computer
  Vision (ECCV)}, pp.~614--630, 2018.

\bibitem{levenberg1944levenbergmarquardt}
K.~Levenberg, ``A method for the solution of certain non-linear problems in
  least squares,'' {\em Quarterly of applied mathematics}, vol.~2, no.~2,
  pp.~164--168, 1944.

\bibitem{roma2002comparative}
N.~Roma, J.~Santos-Victor, and J.~Tom{\'e}, ``A comparative analysis of
  cross-correlation matching algorithms using a pyramidal resolution
  approach,'' in {\em Empirical Evaluation Methods in Computer Vision},
  pp.~117--142, World Scientific, 2002.

\bibitem{wang2020directshape}
R.~Wang, N.~Yang, J.~St{\"u}ckler, and D.~Cremers, ``Directshape: Direct
  photometric alignment of shape priors for visual vehicle pose and shape
  estimation,'' in {\em Proceedings of the 2020 IEEE International Conference
  on Robotics and Automation (ICRA)}, pp.~11067--11073, IEEE, 2020.

\bibitem{stumberg2022dmvio}
L.~Von~Stumberg and D.~Cremers, ``Dm-vio: Delayed marginalization
  visual-inertial odometry,'' {\em IEEE Robotics and Automation Letters}, 2022.

\bibitem{engel2013semi}
J.~Engel, J.~Sturm, and D.~Cremers, ``Semi-dense visual odometry for a
  monocular camera,'' in {\em Proceedings of the IEEE international conference
  on computer vision}, pp.~1449--1456, 2013.

\bibitem{rojas2003kalmanfilter}
R.~Rojas, ``The kalman filter,'' {\em available on www. robocup. mi. fu-berlin.
  de/buch/kalman. pdf}, pp.~1--7, 2003.

\bibitem{markovic2016wrapping}
I.~Markovi{\'c}, J.~{\'C}esi{\'c}, and I.~Petrovi{\'c}, ``On wrapping the
  kalman filter and estimating with the so (2) group,'' in {\em Proceedings of
  the 2016 19th International Conference on Information Fusion (FUSION)},
  pp.~2245--2250, IEEE, 2016.

\bibitem{voigtlaender2019trackrcnn}
P.~Voigtlaender, M.~Krause, A.~Osep, J.~Luiten, B.~B.~G. Sekar, A.~Geiger, and
  B.~Leibe, ``Mots: Multi-object tracking and segmentation,'' in {\em
  Proceedings of the IEEE/CVF Conference on Computer Vision and Pattern
  Recognition}, pp.~7942--7951, 2019.

\bibitem{ren2017rrc}
J.~Ren, X.~Chen, J.~Liu, W.~Sun, J.~Pang, Q.~Yan, Y.-W. Tai, and L.~Xu,
  ``Accurate single stage detector using recurrent rolling convolution,'' in
  {\em Proceedings of the IEEE conference on computer vision and pattern
  recognition}, pp.~5420--5428, 2017.

\bibitem{luiten2018premvos}
J.~Luiten, P.~Voigtlaender, and B.~Leibe, ``Premvos: Proposal-generation,
  refinement and merging for video object segmentation,'' in {\em Asian
  Conference on Computer Vision}, pp.~565--580, Springer, 2018.

\bibitem{luiten2021hota}
J.~Luiten, A.~Osep, P.~Dendorfer, P.~Torr, A.~Geiger, L.~Leal-Taix{\'e}, and
  B.~Leibe, ``Hota: A higher order metric for evaluating multi-object
  tracking,'' {\em International journal of computer vision}, vol.~129, no.~2,
  pp.~548--578, 2021.

\bibitem{bernardin2008clearmot}
K.~Bernardin and R.~Stiefelhagen, ``Evaluating multiple object tracking
  performance: the clear mot metrics,'' {\em EURASIP Journal on Image and Video
  Processing}, vol.~2008, pp.~1--10, 2008.

\bibitem{luiten2020trackeval}
A.~H. Jonathon~Luiten, ``Trackeval.''
  https://github.com/JonathonLuiten/TrackEval, 2020.

\bibitem{bhat2021adabins}
S.~F. Bhat, I.~Alhashim, and P.~Wonka, ``Adabins: Depth estimation using
  adaptive bins,'' in {\em Proceedings of the IEEE/CVF Conference on Computer
  Vision and Pattern Recognition}, pp.~4009--4018, 2021.

\bibitem{pang2021simpletrack}
Z.~Pang, Z.~Li, and N.~Wang, ``Simpletrack: Understanding and rethinking 3d
  multi-object tracking,'' {\em arXiv preprint arXiv:2111.09621}, 2021.

\bibitem{Argoverse2}
B.~Wilson, W.~Qi, T.~Agarwal, J.~Lambert, J.~Singh, S.~Khandelwal, B.~Pan,
  R.~Kumar, A.~Hartnett, J.~K. Pontes, D.~Ramanan, P.~Carr, and J.~Hays,
  ``Argoverse 2: Next generation datasets for self-driving perception and
  forecasting,'' in {\em Proceedings of the NeurIPS Datasets and Benchmarks
  2021}, 2021.

\bibitem{barber1996quickhull}
C.~B. Barber, D.~P. Dobkin, and H.~Huhdanpaa, ``The quickhull algorithm for
  convex hulls,'' {\em ACM Transactions on Mathematical Software (TOMS)},
  vol.~22, no.~4, pp.~469--483, 1996.

\end{thebibliography}
